\def\year{2018}\relax
\documentclass[letterpaper]{article} %DO NOT CHANGE THIS
\usepackage{aaai18}  %Required
\usepackage{times}  %Required
\usepackage{helvet}  %Required
\usepackage{courier}  %Required
\usepackage{url}  %Required
\usepackage{graphicx}  %Required

\usepackage{color}
\usepackage{amsmath}
\usepackage{amsfonts}
\usepackage{bbding}

 \usepackage{hyperref}

\begin{document}
\title{Spatial Temporal Graph Convolutional Networks for Skeleton-Based Action Recognition}
\author{Sijie Yan, Yuanjun Xiong, Dahua Lin\\
Department of Information Engineering, The Chinese University of Hong Kong\\
\{ys016, dhlin\}@ie.cuhk.edu.hk, bitxiong@gmail.com\\
}
\maketitle

\begin{abstract}

Dynamics of human body skeletons convey significant information
for human action recognition. Conventional approaches for modeling
skeletons usually rely on hand-crafted parts or traversal rules,
thus resulting in limited expressive power and difficulties
of generalization.
In this work, we propose a novel model of dynamic skeletons
called \emph{Spatial-Temporal Graph Convolutional Networks (ST-GCN)},
which moves beyond the limitations of previous methods by
automatically learning both the spatial and temporal patterns from data.
This formulation not only leads to greater expressive power
but also stronger generalization capability.
On two large datasets,
\emph{Kinetics} and \emph{NTU-RGBD}, it achieves substantial
improvements over mainstream methods.

\end{abstract}

\section{Introduction}

Human action recognition has become an active research area in recent years,
as it plays a significant role in video understanding.
In general, human action can be recognized from multiple modalities\cite{Simonyan2014NIPS,Tran2015C3D,Wang2015action,TSN2016ECCV,SSN2017ICCV},
such as appearance, depth, optical flows, and body skeletons~\cite{Du2015CVPR,Liu2016ECCV}.
Among these modalities, \emph{dynamic human skeletons} usually convey
significant information that is complementary to others.
However, the modeling of dynamic skeletons has received relatively less
attention than that of appearance and optical flows.
In this work, we systematically study this modality, with an aim to
develop a principled and effective method to model dynamic skeletons
and leverage them for action recognition.

\begin{figure}
\centering
\includegraphics[width=0.6\linewidth]{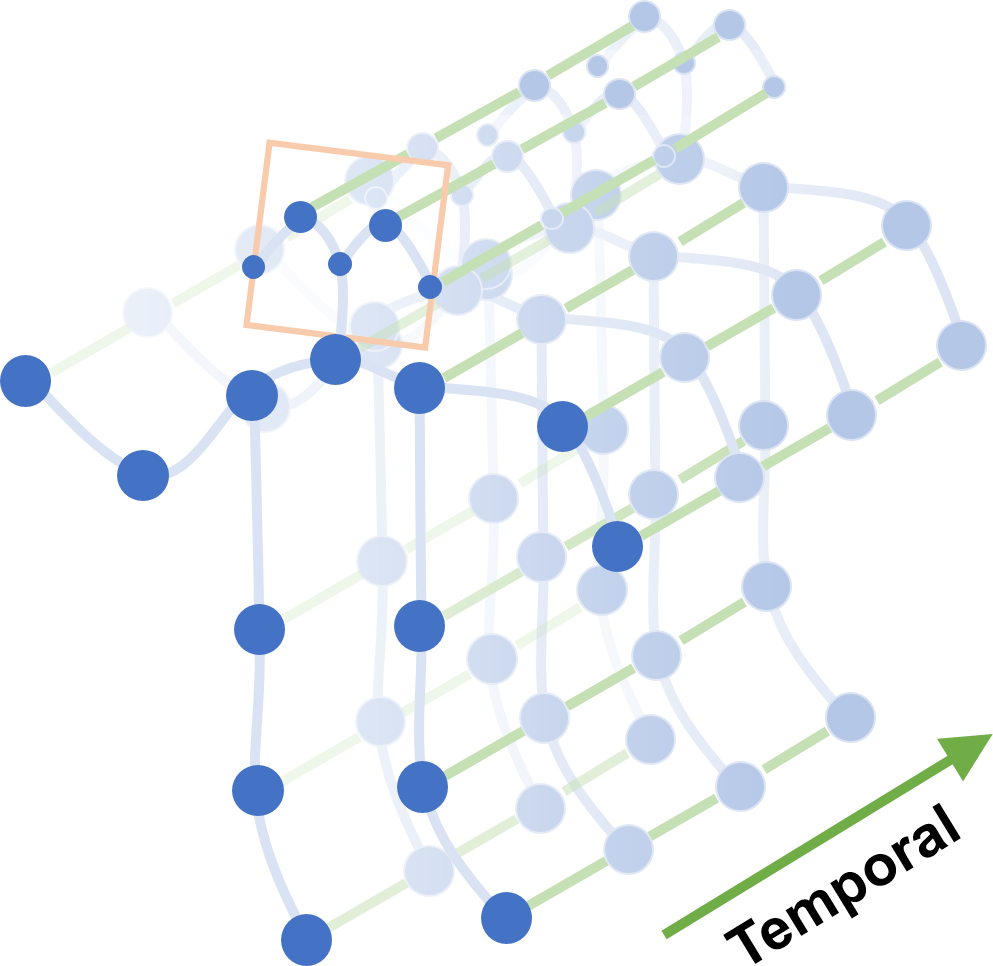}
\caption{\small
	The spatial temporal graph of a skeleton sequence used
	in this work where the proposed ST-GCN operate on.
	Blue dots denote the body joints.
	The intra-body edges between body joints are defined
	based on the natural connections in human bodies.
	The inter-frame edges connect the same joints between consecutive frames.
	Joint coordinates are used as inputs to the ST-GCN.}
\label{fig:teaser_intro}
\end{figure}

The dynamic skeleton modality can be naturally represented
by a time series of human joint locations, in the form of
2D or 3D coordinates. Human actions can then be recognized
by analyzing the motion patterns thereof.
Earlier methods of using skeletons for action recognition
simply employ the joint coordinates at individual time steps
to form feature vectors, and apply temporal analysis thereon~\cite{Wang2012CVPR,Fernando2015VideoDarwin}.
The capability of these methods is limited as they do not explicitly
exploit the spatial relationships among the joints, which are crucial
for understanding human actions.
Recently, new methods that attempt to leverage the natural connections between
joints have been developed~\cite{Shahroudy2016CVPR,Du2015CVPR}. These methods show encouraging
improvement, which suggests the significance of the connectivity.
Yet, most existing methods rely on hand-crafted parts or rules to
analyze the spatial patterns. As a result, the models devised for
a specific application are difficult to be generalized to others.

To move beyond such limitations, we need a new method that can
automatically capture the patterns embedded in the
\emph{spatial configuration} of the joints as well as their
\emph{temporal dynamics}. This is the strength of deep neural networks.
However, as mentioned, the skeletons are in the form of graphs
instead of a 2D or 3D grids, which makes it difficult to use proven
models like convolutional networks.
Recently, \emph{Graph Neural networks (GCNs)}, which generalize
\emph{convolutional neural networks (CNNs)} to graphs of arbitrary structures,
have received increasing attention and successfully been adopted in
a number of applications, such as
image classification~\cite{Bruna2014ICLR},
document classification~\cite{Defferrard2016NIPS},
and semi-supervised learning~\cite{Kipf2017ICLR}.
However, much of the prior work along this line assumes a fixed graph
as input. The application of \emph{GCNs} to model dynamic graphs
over large-scale datasets, \textit{e.g.}~human skeleton sequences, is
yet to be explored.

In this paper, we propose to design a generic representation of skeleton
sequences for action recognition by extending graph neural networks
to a spatial-temporal graph model,
called \emph{Spatial-Temporal Graph Convolutional Networks (ST-GCN)}.
As illustrated in Figure~\ref{fig:teaser_intro}
this model is formulated on top of a sequence of skeleton graphs,
where each node corresponds to a joint of the human body. There are
two types of edges, namely the \emph{spatial edges} that conform to the
natural connectivity of joints and the \emph{temporal edges} that
connect the same joints across consecutive time steps.
Multiple layers of spatial temporal graph convolution
are constructed thereon, which allow information to be integrated
along both the spatial and the temporal dimension.

The hierarchical nature of \emph{ST-GCN} eliminates the need of hand-crafted
part assignment or traversal rules. This not only leads to greater
expressive power and thus higher performance (as shown in our experiments),
but also makes it easy to generalize to different contexts.
Upon the generic GCN formulation, we also study new strategies to
design graph convolution kernels, with inspirations from image models.

The major contributions of this work lie in three aspects:
1) We propose \emph{ST-GCN}, a generic graph-based formulation
for modeling dynamic skeletons, which is the first that applies graph-based
neural networks for this task.
2) We propose several principles in designing convolution kernels in ST-GCN
to meet the specific demands in skeleton modeling.
3) On two large scale datasets for skeleton-based action
recognition, the proposed model achieves superior performance
as compared to previous methods using hand-crafted parts or traversal rules, with considerably less effort in manual design.
The code and models of ST-GCN are made publicly available\footnote{\url{https://github.com/yysijie/st-gcn}}.

\section{Related work}

\paragraph{Neural Networks on Graphs.}
Generalizing neural networks to data with graph structures is an emerging topic in deep learning research.
The discussed neural network architectures include both recurrent neural networks~\cite{Tai2015ACL,Van2016Pixel} and convolutional neural networks (CNNs)~\cite{Bruna2014ICLR,Henaff2015Arxiv,Duvenaud2015NIPS,Li2016ICLR,Defferrard2016NIPS}.
This work is more related to the generalization of CNNs, or graph convolutional networks (GCNs).
The principle of constructing GCNs on graph generally follows two streams: 1) the \textbf{spectral perspective}, where the locality of the graph convolution is considered in the form of spectral analysis~\cite{Henaff2015Arxiv,Duvenaud2015NIPS,Li2016ICLR,Kipf2017ICLR}; 2) the \textbf{spatial perspective}, where the convolution filters are applied directly on the graph nodes and their neighbors~\cite{Bruna2014ICLR,Niepert2016ICML}.
This work follows the spirit of the second stream. We construct the CNN filters on the spatial domain, by limiting the application of each filter to the 1-neighbor of each node.

\paragraph{Skeleton Based Action Recognition.}
Skeleton and joint trajectories of human bodies are robust to illumination change and scene variation, and they are easy to obtain owing to the highly accurate depth sensors or pose estimation algorithms~\cite{Shotton2013CVPR,Cao2017CVPR}. There is thus a broad array of skeleton based action recognition approaches.
The approaches can be categorized into \textbf{hand-crafted feature based methods} and \textbf{deep learning methods}.
The first type of approaches design several handcrafted features to capture the dynamics of joint motion. These could be covariance matrices of joint trajectories~\cite{Hussein2013IJCAI}, relative positions of joints~\cite{Wang2012CVPR}, or rotations and translations between body parts~\cite{Vemulapalli2014CVPR}.
The recent success of deep learning has lead to the surge of deep learning based skeleton modeling methods. These works have been using recurrent neural networks~\cite{Shahroudy2016CVPR,Zhu2016AAAI,Liu2016ECCV,Zhang2017WACV} and temporal CNNs~\cite{Li2017Skeleton,Ke2017CVPR,Kim2017CVPRW} to learn action recognition models in an end-to-end manner.
Among these approaches, many have emphasized the importance of modeling the joints within parts of human bodies.
But these parts are usually explicitly assigned using domain knowledge.
Our ST-GCN is the first to apply graph CNNs to the task of skeleton based action recognition. It differentiates from previous approaches in that it can learn the part information implicitly by harnessing locality of graph convolution together with the temporal dynamics.
By eliminating the need for manual part assignment, the model is easier to design and potent to learn better action representations.

\section{Spatial Temporal Graph ConvNet}
When performing activities, human joints move in small local groups, known as ``body parts''.
Existing approaches for skeleton based action recognition have verified the effectiveness of introducing body parts in the modeling~\cite{Shahroudy2016CVPR,Liu2016ECCV,Zhang2017WACV}. 
We argue that the improvement is largely due to that parts restrict the modeling of joints trajectories within ``local regions'' compared with the whole skeleton, thus forming a hierarchical representation of the skeleton sequences.
In tasks such as image object recognition, the hierarchical representation and locality are usually achieved by the intrinsic properties of  convolutional neural networks~\cite{Krizhevsky2012Imagenet}, rather than manually assigning object parts.
It motivates us to introduce the appealing property of CNNs to skeleton based action recognition.
The result of this attempt is the ST-GCN model.
\begin{figure*}
	\centering
	\includegraphics[width=1.0\linewidth]{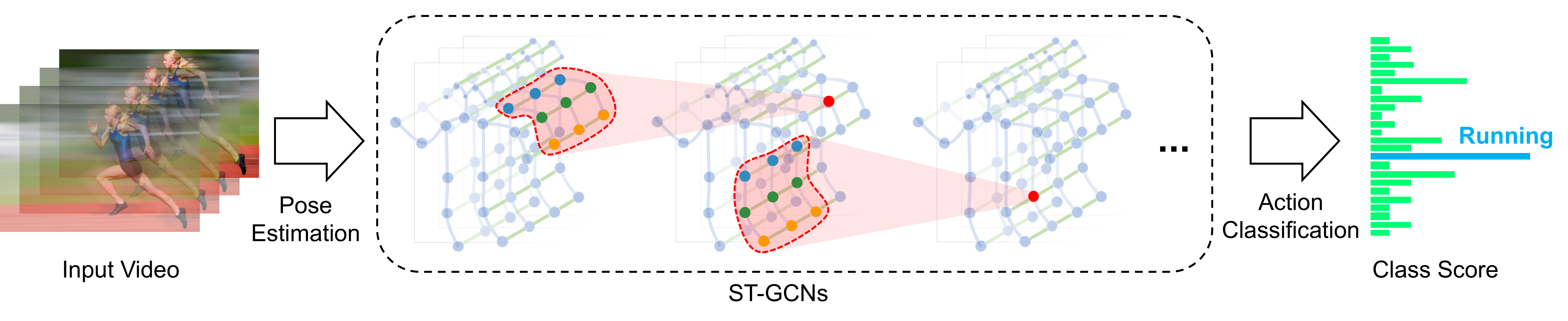}
	\caption{
		We perform pose estimation on videos and construct spatial temporal graph on skeleton sequences. Multiple layers
		of spatial-temporal graph convolution (ST-GCN) will be applied and gradually generate higher-level feature maps on the graph.
		It will then be classified by the standard Softmax classifier to the corresponding action category.
	}
	\label{fig:pipeline}
\end{figure*}
\subsection{Pipeline Overview}
Skeleton based data can be obtained from motion-capture devices or pose estimation algorithms from videos.
Usually the data is a sequence of frames, each frame will have a set of joint coordinates.
Given the sequences of body joints in the form of 2D or 3D coordinates, we construct a spatial temporal graph with the joints as graph nodes and natural connectivities in both human body structures and time as graph edges.
The input to the ST-GCN is therefore the joint coordinate vectors on the graph nodes. 
This can be considered as an analog to image based CNNs where the input is formed by pixel intensity vectors residing on the 2D image grid.
Multiple layers of spatial-temporal graph convolution operations will be applied on the input data and generating higher-level feature maps on the graph.
It will then be classified by the standard SoftMax classifier to the corresponding action category.
The whole model is trained in an end-to-end manner with backpropagation.
We will now go over the components in the ST-GCN model.

\subsection{Skeleton Graph Construction}
\label{sec:graph_construction}
A skeleton sequence is usually represented by 2D or 3D coordinates of each human joint in each frame. Previous work using convolution for skeleton action recognition~\cite{Kim2017CVPRW} concatenates coordinate vectors of all joints to form a single feature vector per frame.
In our work, we utilize the spatial temporal graph to form hierarchical representation of the skeleton sequences. 
Particularly, we construct an undirected spatial temporal graph $ G = (V, E) $ on a skeleton sequence with $ N $ joints and $ T $ frames featuring both intra-body and inter-frame connection.

In this graph, the node set $ V = \{v_{ti} | t = 1,\ldots, T, i=1,\ldots,N\} $ includes the all the joints in a skeleton sequence. 
As ST-GCN's input, the feature vector on a node $F(v_{ti})$ consists of coordinate vectors, as well as estimation confidence, of the $i$-th joint on frame $t$.
We construct the spatial temporal graph on the skeleton sequences in two steps. 
First, the joints within one frame are connected with edges according to the connectivity of human body structure, which is illustrated in Fig.~\ref{fig:teaser_intro}. 
Then each joint will be connected to the same joint in the consecutive frame.
The connections in this setup are thus naturally defined without the manual part assignment. 
This also enables the network architecture to work on datasets with different number of joints or joint connectivities.
For example, on the Kinetics dataset, we use the 2D pose estimation results from the OpenPose~\cite{Cao2017Openpose} toolbox which outputs 18 joints, while on the NTU-RGB+D dataset~\cite{Shahroudy2016CVPR} we use 3D joint tracking results as input, which produces 25 joints.
The ST-GCN can operate in both situations and provide consistent superior performance.
An example of the constructed spatial temporal graph is illustrated in Fig.~\ref{fig:teaser_intro}.

Formally,
the edge set $ E $ is composed of two subsets, the first subset depicts the intra-skeleton connection at each frame, denoted as
$	
E_S = \{v_{ti}v_{tj}| (i, j) \in H\} 
$,
where $ H $ is the set of naturally connected human body joints.
The second subset contains the inter-frame edges, which connect the same joints in consecutive frames as
$
	E_F = \{v_{ti}v_{(t+1) i}\}
$. 
Therefore all edges in $ E_F $ for one particular joint $ i $ will represent its trajectory over time.

\subsection{Spatial Graph Convolutional Neural Network}
Before we dive into the full-fledged ST-GCN, we first look at the graph CNN model within one single frame.
In this case, on a single frame at time $ \tau $, there will be N joint nodes $ V_t $, along with the skeleton edges $ E_S(\tau) = \{v_{ti}v_{tj}| t= \tau, (i, j) \in H\} $.
Recall the definition of convolution operation on the 2D natural images or feature maps, which can be both treated as 2D grids.
The output feature map of a convolution operation is again a 2D grid.
With stride $ 1 $ and appropriate padding, the output feature maps can have the same size as the input feature maps. 
We will assume this condition in the following discussion.
Given a convolution operator with the kernel size of $ K\times K $, and an input feature map $ f_{in} $ with the number of channels $ c $.
The output value for a single channel at the spatial location $ \mathbf{x} $ can be written as 

\begin{equation}\label{eq:gen_conv}
f_{out}(\mathbf{x}) = \sum_{h=1}^{K} \sum_{w=1}^{K} f_{in}(\mathbf{p}(\mathbf{x}, h, w))\cdot \mathbf{w}(h, w), 	
\end{equation}

where the \textbf{sampling function} $ \mathbf{p}: Z^2\times Z^2\rightarrow Z^2 $ enumerates the neighbors of location $ \mathbf{x} $.
In the case of image convolution, it can also be represented as $ \mathbf{p}(\mathbf{x}, h, w) = \mathbf{x} + \mathbf{p^\prime}(h, w). $ 
The \textbf{weight function} $ \mathbf{w}: Z^2\rightarrow \mathbb{R}^c$ provides a weight vector in $ c $-dimension real space for computing the inner product with the sampled input feature vectors of dimension $ c $.
Note that the weight function is irrelevant to the input location $ \mathbf{x} $.
Thus the filter weights are shared everywhere on the input image.
Standard convolution on the image domain is therefore achieved by encoding a rectangular grid in $ \mathbf{p}(\mathbf{x}) $.
More detailed explanation and other applications of this formulation can be found in~\cite{Dai2017Arxiv}.

The convolution operation on graphs is then defined by extending the formulation above to the cases where the input features map resides on a spatial graph $ V_t $.
That is, the feature map $ f_{in}^t : V_t\rightarrow R^c$ has a vector on each node of the graph.
The next step of the extension is to redefine the sampling function $ \mathbf{p} $ and the weight function $ \mathbf{w} $.

\paragraph{Sampling function.} 
On images, the sampling function $\mathbf{p}(h, w)$ is defined on the neighboring pixels with respect to the center location $ \mathbf{x}$. 
On graphs, we can similarly define the sampling function on the neighbor set $ B(v_{ti})=\{v_{tj} | d(v_{tj}, v_{ti}) \leq D \} $ of a node $ v_{ti} $.
Here $ d(v_{tj}, v_{ti}) $ denotes the minimum length of any path from $v_{tj} $ to $ v_{ti} $. 
Thus the sampling function $ \mathbf{p}:B(v_{ti})\rightarrow V $ can be written as
\begin{align}~\label{eq:sampling}
\mathbf{p}(v_{ti}, v_{tj}) = v_{tj}.
\end{align}
In this work we use $ D = 1 $ for all cases, that is, the $ 1 $-neighbor set of joint nodes.
The higher number of $ D $ is left for future works.

\paragraph{Weight function.}
Compared with the sampling function, the weight function is trickier to define.
In 2D convolution, a rigid grid naturally exists around the center location. 
So pixels within the neighbor can have a fixed spatial order. 
The weight function can then be implemented by indexing a tensor of $ (c, K, K) $ dimensions according to the spatial order.
For general graphs like the one we just constructed, there is no such implicit arrangement.
The solution to this problem is first investigated in~\cite{Niepert2016ICML}, where the order is defined by a graph labeling process in the neighbor graph around the root node.
We follow this idea to construct our weight function.
Instead of giving every neighbor node a unique labeling, we simplify the process by partitioning the neighbor set $B(v_{ti})$ of a joint node $ v_{ti} $ into a fixed number of $ K $ subsets, where each subset has a numeric label.
Thus we can have a mapping $ l_{ti}:B(v_{ti})\rightarrow \{0,\ldots,K-1\} $ which maps a node in the neighborhood to its subset label.
The weight function $ \mathbf{w}(v_{ti}, v_{tj}): B(v_{ti})\rightarrow R^c $ can be implemented by indexing a tensor of $ (c, K) $ dimension or
\begin{align}\label{eq:weight}
\mathbf{w}(v_{ti}, v_{tj}) = \mathbf{w}^\prime(l_{ti}(v_{tj})).
\end{align}
We will discuss several partitioning strategies in Sec.~\ref{sec:partition}.

\paragraph{Spatial Graph Convolution.}
With the refined sampling function and weight function, we now rewrite Eq.~\ref{eq:gen_conv} in terms of graph convolution as
\begin{align}
	\label{eq:graph_conv}
	f_{out}(v_{ti}) = \sum_{v_{tj}\in B(v_{ti})} \frac{1}{Z_{ti}(v_{tj})}f_{in}(\mathbf{p}(v_{ti}, v_{tj}))\cdot \mathbf{w}(v_{ti}, v_{tj}), 
\end{align}
where the normalizing term 
$
Z_{ti}(v_{tj}) = \mid\{ v_{tk} |  l_{ti}(v_{tk}) = l_{ti}(v_{tj})\}\mid
$
equals the cardinality of the corresponding subset.
This term is added to balance the contributions of different subsets to the output.
Substituting Eq.~\ref{eq:sampling} and Eq.~\ref{eq:weight} into Eq.~\ref{eq:graph_conv}, we arrive at
\begin{align}
		\label{eq:graph_conv_simplify}
		f_{out}(v_{ti}) = \sum_{v_{tj}\in B(v_{ti})} \frac{1}{Z_{ti}(v_{tj})}f_{in}(v_{tj})\cdot \mathbf{w}(l_{ti}(v_{tj})).
\end{align}
It is worth noting this formulation can resemble the standard 2D convolution if we treat a image as a regular 2D grid.
For example, to resemble a $ 3\times 3 $ convolution operation, we have a neighbor of $ 9 $ pixels in the $ 3\times 3 $ grid centered on a pixel. 
The neighbor set should then be partitioned into $ 9 $ subsets, each having one pixel.

\paragraph{Spatial Temporal Modeling.}
Having formulated spatial graph CNN, we now advance to the task of modeling the spatial temporal dynamics within skeleton sequence.
Recall that in the construction of the graph, the temporal aspect of the graph is constructed by connecting the same joints across consecutive frames.
This enable us to define a very simple strategy to extend the spatial graph CNN to the spatial temporal domain. 
That is, we extend the concept of neighborhood to also include temporally connected joints as
\begin{align}\label{eq:neighbor_temp}
	 B(v_{ti})=\{v_{qj} | d(v_{tj}, v_{ti}) \leq K, |q - t| \leq \lfloor\Gamma/2\rfloor \}.
\end{align}
The parameter $ \Gamma $ controls the temporal range to be included in the neighbor graph and can thus be called the temporal kernel size.
To complete the convolution operation on the spatial temporal graph, we also need the sampling function, which is the same as the spatial only case, and the weight function, or in particular, the labeling map $ l_{ST} $. 
Because the temporal axis is well-ordered, we directly modify the label map $ l_{ST} $ for a spatial temporal neighborhood rooted at $ v_{ti}$ to be
\begin{align}\label{eq:spatial_to_temporal}
	l_{ST}(v_{qj}) = l_{ti}(v_{tj}) + (q - t + \lfloor\Gamma/2\rfloor) \times K,
\end{align}
where $ l_{ti}(v_{tj}) $ is the label map for the single frame case at $ v_{ti} $.
In this way, we have a well-defined convolution operation on the constructed spatial temporal graphs.

\begin{figure*}
	\centering
	\includegraphics[width=0.6\linewidth]{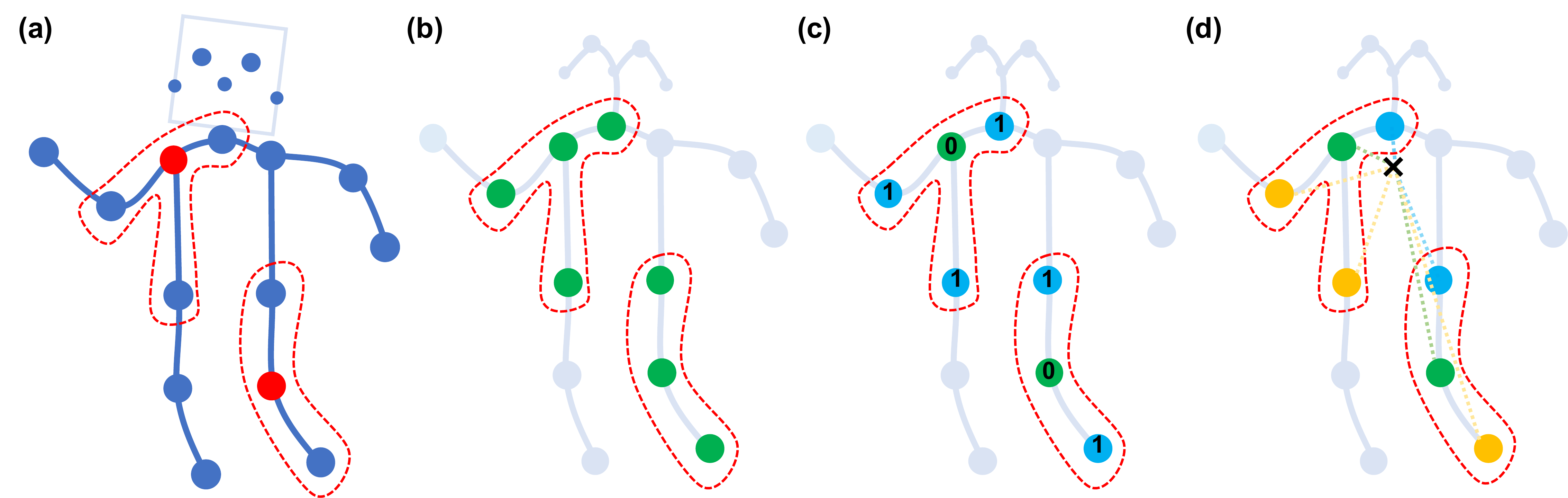}
	\caption{The proposed partitioning strategies for constructing convolution operations. 
		From left to right: 
		\textbf{(a)} An example frame of input skeleton. Body joints are drawn with blue dots. The receptive fields of a filter with $ D=1 $ are drawn with red dashed circles. 
		\textbf{(b)} \textbf{Uni-labeling} partitioning strategy, where all nodes in a neighborhood has the same label (green).
		\textbf{(c)} \textbf{Distance} partitioning. The two subsets are the root node itself with distance $0$ (green) and other neighboring points with distance $1$. (blue).
		\textbf{(d)} \textbf{Spatial configuration} partitioning. The nodes are labeled according to their distances to the skeleton gravity center (black cross) compared with that of the root node (green).
		Centripetal nodes have shorter distances (blue), while centrifugal nodes have longer distances (yellow) than the root node.
	}
	\label{fig:partition}
\end{figure*}

\subsection{Partition Strategies.}\label{sec:partition}
Given the high-level formulation of spatial temporal graph convolution, it is important to design a partitioning strategy to implement the label map $ l $.
In this work we explore several partition strategies.
For simplicity, we only discuss the cases in a single frame because they can be naturally extended to the spatial-temporal domain using Eq.~\ref{eq:spatial_to_temporal}.

\paragraph{Uni-labeling.}
The simplest and most straight forward partition strategy is to have subset, which is the whole neighbor set itself.
In this strategy, feature vectors on every neighboring node will have a inner product with the same weight vector. 
Actually, this strategy resembles the propagation rule introduced in~\cite{Kipf2017ICLR}.
It has an obvious drawback that in the single frame case, using this strategy is equivalent to computing the inner product between the weight vector and the average feature vector of all neighboring nodes. 
This is suboptimal for skeleton sequence classification as the local differential properties could be lost in this operation.
Formally, we have $K=1$ and $l_{ti}(v_{tj}) = 0, \forall i,j \in V$.

\paragraph{Distance partitioning.}
Another natural partitioning strategy is to partition the neighbor set according to the nodes' distance $ d(\cdot, v_{ti}) $ to the root node $ v_{ti} $.
In this work, because we set $ D = 1 $, the neighbor set will then be separated into two subsets, where $ d = 0 $ refers to the root node itself and remaining neighbor nodes are in the $d=1$ subset.
Thus we will have two different weight vectors and they are capable of modeling local differential properties such as the relative translation between joints. Formally, we have $K=2$ and
$l_{ti}(v_{tj}) = d(v_{tj}, v_{ti})$ .

\paragraph{Spatial configuration partitioning.}
Since the body skeleton is spatially localized, we can still utilize this specific spatial configuration in the partitioning process.
We design a strategy to divide the neighbor set into three subsets:
1) the root node itself; 2)centripetal group: the neighboring nodes that are closer to the gravity center of the skeleton than the root node; 3) otherwise the centrifugal group.
Here the average coordinate of all joints in the skeleton at a frame is treated as its gravity center.
This strategy is inspired by the fact that motions of body parts can be broadly categorized as concentric and eccentric motions. Formally, we have 
\begin{equation}
l_{ti}({v_tj})=
	\begin{cases}
	   0 &\mbox{if $r_{j} = r_{i}$}\\
	   1 &\mbox{if $r_{j} < r_{i}$}\\
	   2 &\mbox{if $r_{j} > r_{i}$}
	\end{cases}
\end{equation}
where $r_i$ is the average distance from gravity center to joint $i$ over all frames in the training set.
 
Visualization of the three partitioning strategies is shown in Fig.~\ref{fig:partition}. We will empirically examine the proposed partioning strategies on skeleton based action recognition experiments. 
It is expected that a more advanced partitioning strategy will lead to better modeling capacity and recognition performance.

\subsection{Learnable edge importance weighting.}
Although joints move in groups when people are performing actions, one joint could appear in multiple body parts. These appearances, however, should have different importance in modeling the dynamics of these parts. 
In this sense, we add a learnable mask $ \mathbf{M} $ on every layer of spatial temporal graph convolution. 
The mask will scale the contribution of a node's feature to its neighboring nodes based on the learned importance weight of each spatial graph edge in $ E_S $.
Empirically we find adding this mask can further improve the recognition performance of ST-GCN.
It is also possible to have a data dependent attention map for this sake. We leave this to future works.

\subsection{Implementing ST-GCN}
The implementation of graph-based convolution is not as straightforward as 2D or 3D convolution. 
Here we provide details on implementing ST-GCN for skeleton based action recognition.

We adopt a similar implementation of graph convolution as in~\cite{Kipf2017ICLR}.
The intra-body connections of joints within a single frame are represented by an adjacency matrix $ \mathbf{A} $ and an identity matrix $ \mathbf{I} $ representing self-connections.
In the single frame case, ST-GCN with the first partitioning strategy can be implemented with the following formula~\cite{Kipf2017ICLR}
\begin{align}\label{eq:graph_conv_uni}
	\mathbf{f}_{out} = \mathbf{\Lambda}^{-\frac{1}{2}}(\mathbf{A} + \mathbf{I})\mathbf{\Lambda}^{-\frac{1}{2}}\mathbf{f}_{in}\mathbf{W},
\end{align}
where $ \Lambda^{ii} = \sum_j(A^{ij} + I^{ij}) $.
Here the weight vectors of multiple output channels are stacked to form the weight matrix $ \mathbf{W} $.
In practice, under the spatial temporal cases, we can represent the input feature map as a tensor of $(C, V, T)$ dimensions.
The graph convolution is implemented by performing a $ 1\times \Gamma $ standard 2D convolution and multiplies the resulting tensor with the normalized adjacency matrix $ \mathbf{\Lambda}^{-\frac{1}{2}}(\mathbf{A} + \mathbf{I})\mathbf{\Lambda}^{-\frac{1}{2}} $ on the second dimension.

For partitioning strategies with multiple subsets, \emph{i.e.}, distance partitioning and spatial configuration partitioning, we again utilize this implementation. 
But note now the adjacency matrix is dismantled into several matrixes $ \mathbf{A}_j $ where $ \mathbf{A} + \mathbf{I} = \sum_j \mathbf{A}_j $.
For example in the distance partitioning strategy, $ \mathbf{A}_0 = \mathbf{I} $ and $ \mathbf{A}_1 = \mathbf{\mathbf{A}} $. 
The Eq.~\ref{eq:graph_conv_uni} is transformed into
\begin{align}\label{eq:graph_conv_part}
	\mathbf{f}_{out} = \sum_j \mathbf{\Lambda}_j^{-\frac{1}{2}}\mathbf{A}_j\mathbf{\Lambda}_j^{-\frac{1}{2}}\mathbf{f}_{in}\mathbf{W}_j,
\end{align}
where similarly $ \Lambda^{ii}_j = \sum_k(A^{ik}_j) + \alpha$. 
Here we set $ \alpha = 0.001 $ to avoid empty rows in $ \mathbf{A}_j $.

It is straightforward to implement the learnable edge importance weighting. 
For each adjacency matrix, we accompany it with a learnable weight matrix $ \mathbf{M} $.
And we substitute the matrix $ \mathbf{A} + \mathbf{I} $ in Eq.~\ref{eq:graph_conv_uni} and $ {A}_j $ in $ \mathbf{A}_j $ in Eq.~\ref{eq:graph_conv_part} with $(\mathbf{A} + \mathbf{I})\otimes\mathbf{M}$ and $ \mathbf{A}_j\otimes \mathbf{M} $, respectively. 
Here $ \otimes $ denotes element-wise product between two matrixes.
The mask $ \mathbf{M} $ is initialized as an all-one matrix. 

\paragraph{Network architecture and training.}
Since the ST-GCN share weights on different nodes, it is important to keep the scale of input data consistent on different joints. 
In our experiments, we first feed input skeletons to a batch normalization layer to normalize data.   
The ST-GCN model is composed of $9$ layers of spatial temporal graph convolution operators (ST-GCN units). 
The first three layers have $ 64 $ channels for output. 
The follow three layers have $128$ channels for output.
And the last three layers have $256$ channels for output.
These layers have $ 9 $ temporal kernel size.
The Resnet mechanism is applied on each ST-GCN unit.
And we randomly dropout the features at $0.5$ probability after each ST-GCN unit to avoid overfitting. 
The strides of the $4$-th and the $7$-th temporal convolution layers are set to $2$ as
pooling layer.
After that, a global pooling was performed on the resulting tensor to get a $ 256 $ dimension feature vector for each sequence. 
Finally, we feed them to a SoftMax classifier.
The models are learned using stochastic gradient descent with a learning rate of $ 0.01 $. We decay the learning rate by $ 0.1 $ after every $ 10 $ epochs.
To avoid overfitting, we perform two kinds of augmentation to replace dropout layers when training on the Kinetics dataset~\cite{Kay2017Kinetics}.
First, to simulate the camera movement, we perform random affine transformations on the skeleton sequences of all frames. 
Particularly, from  the first frame to the last frame, we select a few fixed angle, translation and scaling factors as candidates 
and then randomly
sampled two combinations of three factors to generate an affine transformation. 
This transformation is interpolated for intermediate frames to generate a effect as if we smoothly move the view point during playback.
We name this augmentation as \emph{random moving}.
Second, we randomly sample fragments from the original skeleton sequences in training and use all frames in the test. 
Global pooling at the top of the network enables the network to handle the input sequences with indefinite length.

\section{Experiments}

In this section we evaluate the performance of ST-GCN in skeleton based action recognition experiments.
We experiment on two large-scale action recognition datasets with vastly different properties: \textbf{Kinetics human action dataset} (Kinetics)~\cite{Kay2017Kinetics} is by far the largest unconstrained action recognition dataset, and \textbf{NTU-RGB+D}~\cite{Shahroudy2016CVPR} the largest in-house captured action recognition dataset.
In particular, we first perform detailed ablation study on the Kinetics dataset to examine the contributions of the proposed model components to the recognition performance.
Then we compare the recognition results of ST-GCN with other state-of-the-art methods and other input modalities.
To verify whether the experience we gained on in the unconstrained setting is universal, we experiment with the constraint setting on NTU-RGB+D and compare ST-GCN with other state-of-the-art approaches.
All experiments were conducted on PyTorch deep learning framework with 8 TITANX GPUs.

\subsection{Dataset \& Evaluation Metrics}
\textbf{Kinetics.}
Deepmind Kinetics human action dataset~\cite{Kay2017Kinetics} contains around $300,000$ video clips retrieved from YouTube.
The videos cover as many as $400$ human action classes, ranging from daily activities, sports scenes, to complex actions with interactions.
Each clip in Kinetics lasts around $10$ seconds.

This Kinetics dataset provides only raw video clips without skeleton data. 
In this work we are focusing on skeleton based action recognition, so we use the estimated joint locations in the pixel coordinate system as our input and discard the raw RGB frames.
To obtain the joint locations, we first resize all videos to the resolution of $340 \times 256$ and convert the frame rate to $30$ FPS. Then we use the public available \emph{OpenPose}~\cite{Cao2017Openpose} toolbox to estimate the location of $ 18 $ joints on every frame of the clips.
The toolbox gives 2D coordinates $(X, Y)$ in the pixel coordinate system and confidence scores $C$ for the $18$ human joints.
We thus represent each joint with a tuple of $(X, Y, C)$ and a skeleton frame is recorded as an array of $ 18 $ tuples.
For the multi-person cases, we select $2$ people with the highest average joint confidence in each clip.
In this way, one clip with $ T $ frames is transformed into a skeleton sequence of these tuples.
In practice, we represent the clips with tensors of $ (3, T, 18, 2) $ dimensions.
For simplicity, we pad every clip by replaying the sequence from the start to have $ T=300 $.
We will release the estimated joint locations on Kinetics for reproducing the results.

We evaluate the recognition performance by  top-$1 $ and  top-$5 $ classification accuracy as recommended by the dataset authors~\cite{Kay2017Kinetics}. 
The dataset provides a training set of $ 240,000 $ clips and a validation set of $ 20,000 $.
We train the compared models on the training set and report the accuracies on the validation set.

\textbf{NTU-RGB+D:}
NTU-RGB+D~\cite{Shahroudy2016CVPR} is currently the largest dataset with 3D joints annotations for human action recognition task.
This dataset contains $ 56,000 $ action clips in $60$ action classes.
These clips are all performed by $ 40 $ volunteers captured in a constrained lab environment, with three camera views recorded simultaneously.
The provided annotations give 3D joint locations $ (X, Y, Z) $ in the camera coordinate system, detected by the Kinect depth sensors. 
There are $ 25 $ joints for each subject in the skeleton sequences.
Each clip is guaranteed to have at most $ 2 $ subjects.

The authors of this dataset recommend two benchmarks: 
1) \textbf{cross-subject} (X-Sub) benchmark with $ 40,320 $ and $ 16,560 $ clips for training and evaluation. In this setting the training clips come from one subset of actors and the models are evaluated on clips from the remaining actors; 
2) \textbf{cross-view}(X-View) benchmark $ 37,920 $ and $ 18,960 $ clips. Training clips in this setting come from the camera views $ 2 $ and $ 3 $, and the evaluation clips are all from the camera view $ 1 $.
We follow this convention and report the top-$1$ recognition accuracy on both benchmarks.

\begin{table}
	\centering
	\begin{tabular}{c|c|c}
		\hline
		& Top-1 & Top-5 \\ \hline
		Baseline TCN			& 	$20.3\%$ & $40.0\%$\\\hline
		Local Convolution	& 	$22.0\%$ &$43.2\%$ \\\hline
		Uni-labeling	& $19.3\%$	& $37.4\%$\\
		Distance partitioning*	& $23.9\%$	&$44.9\%$  \\
		Distance Partitioning	& $29.1\%$& $51.3\%$ \\
		Spatial Configuration	& $29.9\%$& $52.2\%$ \\\hline
		ST-GCN + Imp.					& $\mathbf{30.7\%}$& $\mathbf{52.8\%}$ \\\hline
	\end{tabular}
	\caption{Ablation study on the Kinetics dataset. The ``ST-GCN+Imp.'' is used in comparison with other state-of-the-art methods.
		For meaning of each setting please refer to Sec.\ref{sec:ablation}. }\label{tab:ablation}
\end{table}

\subsection{Ablation Study}\label{sec:ablation}
We examine the effectiveness of the proposed components in ST-GCN in this section by action recognition experiments on the Kinetics dataset~\cite{Kay2017Kinetics}.

\paragraph{Spatial temporal graph convolution.}
First, we evaluate the necessity of using spatial temporal graph convolution operation.
We use a baseline network architecture~\cite{Kim2017CVPRW} where all spatial temporal convolutions are replaced by only temporal convolution. That is, we concatenate all input joint locations to form the input features at each frame $ t $.
The temporal convolution will then operate on this input and convolves over time.
We call this model ``baseline TCN''.
This kind of recognition models is known to work well on constraint dataset such as NTU-RGB+D~\cite{Kim2017CVPRW}.
Seen from Table~\ref{tab:ablation}, models with spatial temporal graph convolution, with reasonable partitioning strategies, consistently outperform the baseline model on Kinetics.
Actually, this temporal convolution is equivalent to spatial temporal graph convolution with unshared weights on a fully connected joint graph.
So the major difference between the baseline model and ST-GCN models are the sparse natural connections and shared weights in convolution operation.
Additionally, we evaluate an intermediate model between the baseline model and ST-GCN, referred as ``local convolution''. 
In this model we use the sparse joint graph as ST-GCN, but use convolution filters with unshared weights.
We believe the better performance of ST-GCN based models could justify the power of the spatial temporal graph convolution in skeleton based action recognition.

\paragraph{Partition strategies}
In this work we present three partitioning strategies: 1) uni-labeling; 2) distance partitioning; and 3) spatial configuration partitioning.
We evaluate the performance of ST-GCN with these partitioning strategies.
The results are summarized in Table~\ref{tab:ablation}.
We observe that partitioning with multiple subsets is generally much better than uni-labeling. 
This is in accordance with the obvious problem of uni-labeling that it is equivalent to simply averaging features before the convolution operation.
Given this observation, we experiment with an intermediate between the distance partitioning and uni-labeling, referred to as ``distance partitioning*''.
In this setting we bind the weights of the two subsets in distance partitioning to be different only by a scaling factor $-1$, or $ \mathbf{w}_0 = -\mathbf{w}_1 $.
This setting still achieves better performance than uni-labeling, which again demonstrate the importance of the partitioning with multiple subsets.
Among multi-subset partitioning strategies, the spatial configuration partitioning achieves better performance.
This corroborates our motivation in designing this strategy, which takes into consideration the concentric and eccentric motion patterns.
Based on these observations, we use the spatial configuration partitioning strategy in the following experiments.

\paragraph{Learnable edge importance weighting.}
Another component in ST-GCN is the learnable edge importance weighting. 
We experiment with adding this component on the ST-GCN model with spatial configuration partitioning.
This is referred to as ``ST-GCN+Imp.'' in Table~\ref{tab:ablation}.
Given the high performing vanilla ST-GCN, this component is still able to raise the recognition performance by more than $ 1 $ percent.
Recall that this component is inspired by the fact that joints in different parts have different importances.
It is verified that the ST-GCN model can now learn to express the joint importance and improve the recognition performance.
Based on this observation, we always use this component with ST-GCN in comparison with other state-of-the-art models.

\subsection{Comparison with State of the Arts}

To verify the performance of ST-GCN in both unconstrained and constraint environment, we perform experiments on Kinetics dataset~\cite{Kay2017Kinetics} and NTU-RGB+D dataset\cite{Shahroudy2016CVPR}, respectively.

\paragraph{Kinetics.}
On Kinetics, we compare with three characteristic approaches for skeleton based action recognition. The first is the feature encoding approach on hand-crafted features~\cite{Fernando2015VideoDarwin}, referred to as ``Feature Encoding'' in Table~\ref{tab:kinetics}. 
We also implemented two deep learning based approaches on Kinetics, \emph{i.e.} Deep LSTM~\cite{Shahroudy2016CVPR} and Temporal ConvNet~\cite{Kim2017CVPRW}.
We compare the approaches' recognition performance in terms of top-1 and top-5 accuracies.
In Table~\ref{tab:kinetics}, ST-GCN is able to outperform previous representative approaches. 
For references, we list the performance of using RGB frames and optical flow for recognition as reported in~\cite{Kay2017Kinetics}.

\begin{table}
	\centering
	\begin{tabular}{c|c|c}
		\hline
		& Top-1 & Top-5 \\ \hline
		RGB\cite{Kay2017Kinetics}				& 	$57.0\%$ &$77.3\%$\\
		Optical Flow~\cite{Kay2017Kinetics}	& 	$49.5\%$ &$71.9\%$ \\\hline
		Feature Enc.~\cite{Fernando2015VideoDarwin}	& $14.9\%$	& $25.8\%$\\
		Deep LSTM~\cite{Shahroudy2016CVPR}		& $16.4\%$	&$35.3\%$  \\
		Temporal Conv.~\cite{Kim2017CVPRW}	& $20.3\%$& $40.0\%$ \\\hline
		ST-GCN 					&  $\mathbf{30.7\%}$& $\mathbf{52.8\%}$ \\\hline
	\end{tabular}
	\caption{Action recognition performance for skeleton based models on the Kinetics dataset.  On top of the table we list the performance of frame based methods. }\label{tab:kinetics}
\end{table}

\begin{table}
	\small
	\centering
	\begin{tabular}{c|c|c}\hline
		& X-Sub& X-View \\ \hline
		Lie Group~\cite{Veeriah2015CVPR}		& $ 50.1 \% $ & $ 52.8 \% $ \\
		H-RNN~\cite{Du2015CVPR}		& $ 59.1 \% $& $ 64.0 \% $ \\
		Deep LSTM~\cite{Shahroudy2016CVPR}		& $ 60.7 \% $& $ 67.3 \% $ \\
		PA-LSTM~\cite{Shahroudy2016CVPR}			& $ 62.9 \% $& $ 70.3 \% $ \\
		ST-LSTM+TS~\cite{Liu2016ECCV}		& $ 69.2 \% $& $ 77.7 \% $ \\
		Temporal Conv~\cite{Kim2017CVPRW}.		& $ 74.3 \% $ & $ 83.1 \% $ \\
		C-CNN + MTLN~\cite{Ke2017CVPR}		& $ 79.6 \% $& $ 84.8 \% $ \\
		\hline
		ST-GCN 			& $\mathbf{81.5\%} $ & $ \mathbf{88.3\%} $ \\\hline
	\end{tabular}
	\caption{Skeleton based action recognition performance on NTU-RGB+D datasets. We report the accuracies on both the cross-subject (X-Sub) and cross-view (X-View) benchmarks.}\label{tab:ntu}
\end{table}

\paragraph{NTU-RGB+D.}
The NTU-RGB+D dataset is captured in a constraint environment, which allows for methods that require well stabilized skeleton sequences to work well.
We also compare our ST-GCN model with the previous state-of-the-art methods on this dataset.
Due to the constraint nature of this dataset, we do not use any data augmentation when training ST-GCN models.
We follow the standard practice in literature to report cross-subject (X-Sub) and cross-view (X-View) recognition performance in terms of top-1 classification accuracies.
The compared methods include Lie Group~\cite{Veeriah2015CVPR}, Hierarchical RNN~\cite{Du2015CVPR}, Deep LSTM~\cite{Shahroudy2016CVPR}, Part-Aware LSTM (PA-LSTM)~\cite{Shahroudy2016CVPR}, Spatial Temporal LSTM with Trust Gates (ST-LSTM+TS)~\cite{Liu2016ECCV}, Temporal Convolutional Neural Networks (Temporal Conv.)~\cite{Kim2017CVPRW}, and Clips CNN + Multi-task learning (C-CNN+MTLN)~\cite{Ke2017CVPR}.
Our ST-GCN model, with rather simple architecture and no data augmentation as used in~\cite{Kim2017CVPRW,Ke2017CVPR}, is able to outperform previous state-of-the-art approaches on this dataset.

\paragraph{Discussion.}
The two datasets in experiments have very different natures. 
On Kinetics the input is 2D skeletons detected with deep neural networks~\cite{Cao2017CVPR}, while on NTU-RGB+D the input is from Kinect depth sensor.
On NTU-RGB+D the cameras are fixed, while on Kinetics the videos are usually shot by hand-held devices, leading to large camera motion.
The fact that the proposed ST-GCN can work well on both datasets demonstrates the effectiveness of the proposed spatial temporal graph convolution operation and the resultant ST-GCN model.

We also notice that on Kinetics the accuracies of skeleton based methods are inferior to video frame based models~\cite{Kay2017Kinetics}. 
We argue that this is due to a lot of action classes in Kinetics requires recognizing the objects and scenes that the actors are interacting with. 
To verify this, we select a subset of $ 30 $ classes strongly related with body motions, named as ``Kinetics-Motion'' and list the mean class accuracies of skeleton and frame based models~\cite{Kay2017Kinetics} on this subset in Table~\ref{tab:kinetics_motion}. 
We can see that on this subset the performance gap is much smaller. 
We also explore using ST-GCN to capture motion information in two-stream style action recognition. 
As shown as in Fig.~\ref{tab:kinetics_ensemble}, our skeleton based model ST-GCN can also provide complementary information to RGB and optical flow models. 
We train the standard TSN~\cite{TSN2016ECCV} models from scratches on Kinetics with RGB and optical flow models. Adding ST-GCN to the RGB model leads to $0.9\%$ increase, even better than optical flows ($0.8\% $).  Combining RGB, optical flow, and ST-GCN further raises the performance to $71.7\%$. 
These results clearly show that the skeletons can provide complementary information when leveraged effectively (e.g. using ST-GCN).

\begin{table}
	\centering
	\begin{tabular}{c|c|c|c}\hline
		Method & RGB CNN & Flow CNN & ST-GCN \\ \hline
		Accuracy & $ 70.4\% $ & $ 72.8\% $ & $72.4\%$ \\ \hline
	\end{tabular}
	\caption{Mean class accuracies on the ``Kinetics Motion'' subset of the Kinetics dataset. This subset contains 
	$ 30 $ action classes in Kinetics which are strongly related to body motions.}\label{tab:kinetics_motion}
\end{table}

\begin{table}
	\small
	\centering
	\begin{tabular}{c|ccc|c}
		  &RGB TSN   &   Flow TSN &   ST-GCN   &Acc(\%) \\
		\hline
		\hline
		 Single&\checkmark &            &            & 70.3\\ 
		 Model&           & \checkmark &            & 51.0\\ 
		 &           &            & \checkmark & 30.7\\ \hline
		 Ensemble&\checkmark &\checkmark  &            & 71.1\\
		 Model&\checkmark &            & \checkmark & 71.2\\ 
		 &\checkmark &\checkmark  & \checkmark & 71.7\\ 
		% \hline
		
	\end{tabular}
	\caption{Class accuracies on the Kinects dataset \textbf{without} ImageNet pretraining. Although our skeleton based model ST-GCN can not achieve the accuracy of the state of the art model performed on RGB and optical flow modalities, it can provide stronger complementary information than optical flow based model.}\label{tab:kinetics_ensemble}
\end{table}

\section{Conclusion}
In this paper, we present a novel model for skeleton based action recognition, the spatial temporal graph convolutional networks (ST-GCN).
The model constructs a set of spatial temporal graph convolutions on the skeleton sequences.
On two challenging large-scale datasets, the proposed ST-GCN outperforms the previous state-of-the-art skeleton based model. 
In addition, ST-GCN can capture motion information in dynamic skeleton sequences which is complementary to RGB modality. The combination of skeleton based model and frame based model further improves the performance in action recognition.
The flexibility of ST-GCN model also opens up many possible directions for future works. 
For example, how to incorporate contextual information, such as scenes, objects, and interactions into ST-GCN becomes a natural question. 

\paragraph{Acknowledgement} This work is partially supported by the Big Data Collaboration Research grant from SenseTime Group (CUHK Agreement No. TS1610626), and the Early Career Scheme (ECS) of Hong Kong (No. 24204215).

{
	\bibliographystyle{aaai}
	\bibliography{gcn}
}

\end{document}